%% file: main.tex
\documentclass{ecai}

\usepackage{latexsym}
\usepackage{amssymb}
\usepackage{amsmath}
\usepackage{amsthm}
\usepackage{booktabs}
\usepackage{enumitem}
\usepackage{graphicx}
\usepackage{color}
\usepackage{float}

\usepackage{bbm}
\usepackage{multirow}
\usepackage{pifont}

\newcommand{\BibTeX}{B\kern-.05em{\sc i\kern-.025em b}\kern-.08em\TeX}

\newcommand{\blue}{}
\renewcommand{\vec}[1]{\mathbf{#1}} %
\newcommand{\sect}{\textsection}
\newcommand{\cmark}{\ding{51}}%

\definecolor{guohaocolor}{rgb}{0, 0.8196, 0}

\begin{document}

\begin{frontmatter}

\paperid{939}

\title{Leveraging Foundation Models for Zero-Shot IoT Sensing}

\author[1]{\fnms{Dinghao}~\snm{Xue}{\let\thefootnote\relax\footnote{{Delft University of Technology, email: d.xue@student.tudelft.nl, \{g.lan, q.song-1\}@tudelft.nl.}}}}
\author[2]{\fnms{Xiaoran}~\snm{Fan}{\let\thefootnote\relax\footnote{{Google, email: vanxf@google.com.}}}}
\author[3]{\fnms{Tao}~\snm{Chen}{\let\thefootnote\relax\footnote{{University of Pittsburgh, email: tac194@pitt.edu.}}}} 
\author[1]{\fnms{Guohao}~\snm{Lan}}
\author[1]{\fnms{Qun}~\snm{Song}\thanks{Corresponding Author.}}

\begin{abstract}
{\blue Deep learning models are increasingly deployed on edge Internet of Things (IoT) devices. However, these models typically operate under supervised conditions and fail to recognize unseen classes different from training. 
} To address this, zero-shot learning (ZSL) aims to classify data of unseen classes with the help of semantic information. 
Foundation models (FMs) trained on web-scale data have shown impressive ZSL capability in natural language processing and visual understanding. However, leveraging FMs' generalized knowledge for zero-shot IoT sensing using signals such as mmWave, IMU, and Wi-Fi has not been fully investigated.
In this work, we align the IoT data embeddings with the semantic embeddings generated by an FM's text encoder for zero-shot IoT sensing. To utilize the physics principles governing the generation of IoT sensor signals to derive more effective prompts for semantic embedding extraction, we propose to use cross-attention to combine a learnable soft prompt that is optimized automatically on training data and an auxiliary hard prompt that encodes domain knowledge of the IoT sensing task. To address the problem of IoT embeddings biasing to seen classes due to the lack of unseen class data during training, we propose using data augmentation to synthesize unseen class IoT data for fine-tuning the IoT feature extractor and embedding projector. 
We evaluate our approach on multiple IoT sensing tasks. Results show that our approach achieves superior open-set detection and generalized zero-shot learning performance compared with various baselines. 
Our code is available at https://github.com/schrodingho/FM\_ZSL\_IoT.
\end{abstract}
\end{frontmatter}

\input{intro}

\input{related}

\input{problem}

\input{method}

\input{eval}

\input{conclude}

\clearpage
\bibliography{refs}
\clearpage
\appendix
\input{appendix}

\end{document}

%% file: intro.tex
\section{Introduction}
\label{sec:intro}

With the advancement of edge hardware accelerators, deep learning is increasingly employed for IoT sensing tasks on edge devices, such as Wi-Fi human sensing~\cite{YANG2023100703}, sound event detection \cite{xie2021zero}, and activity recognition using motion sensor~\cite{tong2021zero}.
However, although deep learning models show excellent performance in classifying samples from a set of \textit{seen} classes included in the training dataset, {\blue identifying and classifying data samples from \textit{unseen} classes using deep models trained under the supervised setting are challenging.}
To address this, an intuitive solution is to include as many classes as possible during training. However, unlike images, text, and audio, which humans can easily interpret, IoT data often lacks readability and requires costly labeling processes. Thus, IoT datasets usually contain a limited number of classes. For example, most inertial measurement unit (IMU)-based activity recognition datasets contain fewer than 20 activity classes \cite{tong2021zero}, while the ImageNet contains 21,814 classes.

Zero-shot learning (ZSL) \cite{pourpanah2022review} is a promising learning paradigm to address the aforementioned challenge. ZSL classifies data from unseen classes with the help of semantic information that transfers knowledge from seen classes to unseen ones. 
Previous studies rely on manually-engineered attributes as semantic information for zero-shot IoT sensing \cite{wang2017zero,ohashi2018attributes}, which are labor-intensive to design and difficult to scale to complex datasets. 
{\blue 
Some works build semantic spaces using word representation models like Word2Vec \cite{matsuki2019characterizing, wu2020multi}, BERT \cite{xie2021zero}, and GloVe \cite{tong2021zero}. The word vectors are automatically generated 
using
large text corpus, e.g., Wikipedia.
However, the text descriptions may contain information unrelated to the target IoT task. In IMU-based activity recognition, the training dataset may contain redundant text about the IMU sensors and lack motion-related information useful for activity classification. Thus, the word vectors may contain task-irrelevant noise, causing a semantic gap between the IoT data and word embeddings.
}
The work in \cite{tong2021zero} constructs visual semantic space using human activity videos for zero-shot IMU-based human activity recognition, which may raise privacy concerns.
In this work, we aim to explore using foundation models, which are considered to have a generalized understanding of the world acquired from diverse and extensive training data, to generate more effective and contextually relevant semantic embeddings for zero-shot IoT sensing.

Foundation models (FMs) are large-scale deep learning models pre-trained on vast data that serve as the foundation for various downstream tasks \cite{zhou2023comprehensive}. FMs trained on extensive text corpora exhibit remarkable generalizability to a broad spectrum of new tasks, e.g., passing exams \cite{achiam2023gpt}, code generation \cite{nijkamp2022codegen}, and language translation \cite{radford2018improving}. 
Large vision-language FMs embed images with language inputs in a joint semantic space using hundreds of millions of image and text pairs, which achieve impressive zero-shot transferability to downstream tasks like image recognition on unseen datasets \cite{radford2021learning,tang2024data}.
{\blue Inspired by this, recent research jointly aligns audio, depth, infrared, and IMU data with the vision \cite{girdhar2023imagebind} and language \cite{zhu2023languagebind} modalities, aiming to extend the zero-shot capability of the vision-language FMs to multiple modalities. These multi-modal FMs show excellent performance in associating unobserved data pairs of existing modalities.}

Recent research aligns IoT sensor signals to textual semantic features generated by FMs for zero-shot IoT sensing. For example, the work in \cite{zhou2023tent} jointly aligns FM's textual embeddings with multiple IoT sensor signals, including video, LiDAR, and mmWave in a unified semantic space. It demonstrates FM's ZSL capability in recognizing unseen class IoT data. However, this work is built upon large quantities of multi-modal data samples where all modalities are presented together, which are expensive to acquire and impractical if new modalities are to be added to the semantic space.
EdgeFM \cite{yang2023edgefm} leverages FMs for zero-shot sensing on resource-limited edge devices. However, EdgeFM only supports the existing modalities of FMs, including video, images, and audio.

This work aims to leverage FMs' generalized knowledge for zero-shot IoT sensing based on mmWave, IMU, and Wi-Fi signals by aligning the IoT data embeddings with the semantic embeddings generated by an FM’s text encoder.
However, connecting IoT sensor signals with semantic embeddings for effective ZSL is non-trivial. First, IoT sensor signals typically follow certain physics principles, which are strong supervision for effective prompt engineering to generate robust semantic embeddings. To address this, we employ cross-attention to combine a learnable soft prompt that is optimized automatically using training data and an auxiliary hard prompt that encodes domain knowledge. Second, given that the training only involves seen class data, the ZSL model is easily biased to seen classes. To address the bias problem, we propose using data augmentation to synthesize unseen class IoT data for fine-tuning our IoT feature extractor and embedding projector. 
Our approach works as follows. 
We apply prompt engineering on class labels and use an FM's text encoder to extract their semantic embeddings as class prototype representations. Meanwhile, we use an IoT feature extractor to extract features from IoT sensor signals followed by an IoT embedding projector to project the features to the semantic space. During model training, we use contrastive learning to align the class prototypes and IoT embeddings. During zero-shot classification, we conduct open-set detection to identify data of unseen classes and use FM to do zero-shot learning. 
We evaluate our approach on multiple datasets including MM-Fi (mmWave, Wi-Fi), USC-HAD (IMU), and PAMAP2 (IMU). Our approach achieves superior performance in open-set detection and generalized zero-shot learning compared with various baselines.
This paper’s contributions are summarized as follows.
\begin{itemize}
    \item To leverage the domain knowledge for zero-shot IoT sensing, we propose using cross-attention to combine a learnable soft prompt and an auxiliary hard prompt for effective prompt engineering. 
    \item To eliminate the problem of unseen class IoT embeddings biasing to seen class embeddings, we employ data augmentation and open-set detection for generalized zero-shot IoT sensing.
    \item We evaluate our approach on multiple IoT datasets with IMU, mmWave, and Wi-Fi data. The results demonstrate that our approach outperforms various baselines in both open-set detection and generalized zero-shot learning.
\end{itemize}

%% file: related.tex
\section{Background and Related Work}
\label{sec:related}

\textbf{Foundation Models}
(FMs) are general deep learning models that are pre-trained on massive amount of data to support various downstream tasks such as chatbot~\cite{radford2018improving,achiam2023gpt} and image recognition \cite{radford2021learning}. FMs are extensively studied in natural
language processing and computer vision \cite{zhou2023comprehensive}. For example, ChatGPT is fine-tuned for conversational tasks from the generative pre-trained transformer-based language foundation models, e.g., GPT-3.5 \cite{brown2020language} and GPT-4 \cite{achiam2023gpt}. CLIP \cite{radford2021learning} is a vision-language foundation model that trains an image encoder and a text encoder jointly aiming to predict the correct image-text pairs. CLIP achieves zero-shot transferability to unseen image recognition tasks after training on 400 million image-text pairs. More recently, FMs are applied to other modalities, including audio, depth, IMU, and infrared~\cite{zhu2023languagebind,girdhar2023imagebind}. These multi-modal FMs use transformer-based encoders to extract embeddings of different modalities. Then, a joint embedding space is learned via contrastive learning that aligns the embeddings of different modalities with the embedding of a ``binding'' modality, i.e., vision or language. The learned joint embeddings can be used for various tasks such as cross-modal retrieval, cross-modal generation, and composing modalities with arithmetic. The multi-modal FMs trained on different cross-modal data pairs, e.g., (image, text) and (image audio), can implicitly associate unobserved data pairs, e.g., (audio, text), which is defined as \textit{emergent zero-shot} classification. Different from the existing works that focus on FMs' zero-shot transferability on unseen datasets \cite{radford2021learning,tang2024data} and unobserved data pairs \cite{girdhar2023imagebind,zhu2023languagebind}, our work aims to investigate the zero-shot capability of FM characterized by the performance of generalizing to unseen object categories in classification tasks, which represents a more practical scenario in IoT sensing tasks.

\textbf{Zero-Shot Learning}
(ZSL) aims to classify data of unseen classes with the help of semantic information containing knowledge about both seen and unseen classes \cite{pourpanah2022review}. Traditional ZSL methods focus on classifying data into unseen classes. A more realistic setting is the {generalized zero-shot learning} (GZSL) that classifies data samples of seen and unseen classes simultaneously. 
GZSL methods can be categorized as \textit{embedding-based} and \textit{generative-based}. Embedding-based GSZL \cite{akata2015label,liu2018generalized} learns a projection function from data feature space to the semantic space. The goal is to map the data embeddings belonging to the same class to the ground-truth label in the semantic space. 
The embedding-based GZSL is easy to implement but is usually biased towards seen classes due to a lack of unseen class data features during training.
Generative-based GZSL \cite{xian2018feature,chen2021free} trains a model to generate synthetic features of unseen class data based on features of seen class data and semantic information of both seen and unseen classes. The generated features of unseen class data can be used to perform supervised learning, where a model is trained to classify data samples of both seen and unseen classes.
The generative-based GZSL alleviates the biasing problem via synthesizing features of unseen classes. However, the generative models are unstable in training and susceptible to model collapse issue.

\begin{figure*}[h]
    \centering
\includegraphics[width=\textwidth]{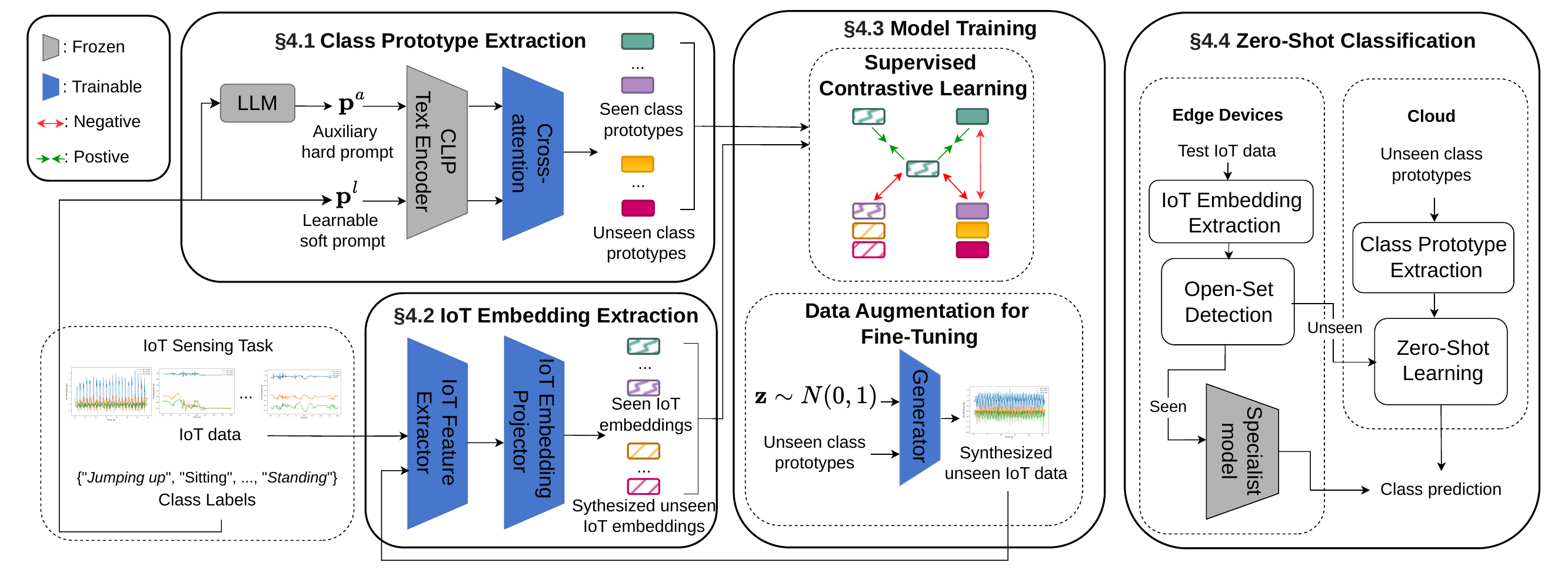}
    \caption{Approach overview. In \sect\ref{sec_4_1}, we use cross-attention to combine the soft and hard prompts to generate class prototypes. In \sect\ref{sec_4_2}, we use a feature extractor followed by an embedding projector to generate IoT embeddings.
    During model training in \sect\ref{sec_4_3}, we use supervised contrastive learning to align the class prototypes and IoT embeddings. We then use data augmentation to synthesize unseen class data for fine-tuning the IoT feature extractor and embedding projector. 
    During zero-shot classification in \sect\ref{sec_4_4}, we first extract the IoT embeddings of input data for open-set detection. Then, the samples detected as seen class will be classified by the specialist model on edge devices. The samples detected as unseen will be uploaded to the cloud for zero-shot classification.
    }
    \label{fig:overview}
    \vspace{1em}
\end{figure*}

\textbf{Zero-Shot IoT Sensing.}
Some works use hand-crafted attributes such as the movement of body or related objects and environment to construct semantic information for zero-shot IoT sensing \cite{wang2017zero,ohashi2018attributes}, which is labour-intensive and less scalable to large complex datasets. To circumvent manual attribute engineering processes, some studies utilize word vectors, which are numerical representations of words in a continuous vector space extracted by word representation models such as Word2Vec~\cite{matsuki2019characterizing, wu2020multi}, BERT~\cite{xie2021zero}, and GloVe~\cite{tong2021zero}, to construct semantic space. The word vectors are extracted 
by capturing the semantic relationships between words based on their contexts in large text corpus.
However, these vectors may include task-irrelevant noise and may not directly suit the specific IoT sensing task.
The work in \cite{tong2021zero} proposes to construct visual semantic space using videos of human activities for IMU-based zero-shot human activity recognition, which is shown to outperform the word vector semantic space. However, collecting videos of human raises privacy concerns.
A recent work \cite{zhou2023tent} jointly aligns multiple IoT data embeddings, including video, LiDAR, and mmWave, with text embeddings extracted from a vision-language FM, CLIP \cite{radford2021learning}, for human activity recognition. With the unified semantic space, not only actions of seen classes can be identified but also the actions of unseen classes can be recognized by the closet textual embedding in the semantic space. However, this approach requires joint training on a self-collected multi-modal aligned dataset, which has limited usability in reality if additional sensor modalities are to be added to the system.
EdgeFM \cite{yang2023edgefm} is an edge-cloud cooperative system that achieves zero-shot recognition capability on resource-limited edge devices by leveraging FMs on the cloud for selective knowledge query. However, the zero-shot capability is only demonstrated on the existing modalities of FMs, including video, images, and audio. 
To this end, the potential of leveraging FMs' generalized knowledge for zero-shot sensing using IoT signals such as mmWave, IMU, and Wi-Fi, which are not covered by the supported modalities of existing FMs, is still under-explored.

%% file: problem.tex
\section{Problem Formulation}
\label{sec:problem-formulation}
We target a deep learning-based IoT sensing task enabled by an edge-cloud cooperative system that contains the following components.
\begin{itemize}
    \item \textbf{Edge Devices} host a small-scale \textit{specialist} deep neural network (DNN) $f(\cdot)$, which can classify a limited set of seen classes $\mathcal{S} = \{c_i^s\}_{i = 1}^{N_s}$. The $f(\cdot)$ is trained under supervised setting using a seen train set $D^s = \{(\vec{x}^s_i, y^s_i)\}^{N_{train}}_{i = 1} \in \mathcal{X} \times \mathcal{S}$, where $\vec{x}^s_i \in \mathbb{R}^d$ is the raw IoT data, $y^s_i$ is the ground-truth label, and $\mathcal{X}$ denotes the IoT data space. The input test data may include not only samples from known seen classes but also samples from novel unseen classes, denoted by 
    $D^{test} = \{\vec{x}^\text{test}_i\}^{N_{test}}_{i = 1} \in \mathcal{X}$.
    \item \textbf{Cloud Server} runs a large \textit{foundation} model (FM) $\Phi(\cdot)$, which possesses general knowledge learned from web-scale training data and has the potential of zero-shot classification on unseen class data. The cloud maintains a list of interested unseen classes outside the set of seen classes $S$, denoted by $\mathcal{U} = \{c_i^u\}_{i = 1}^{N_u}$, where $\mathcal{S}\cap \mathcal{U} = \emptyset$. Note that $\mathcal{U}$ can be specified by users or include the commonly seen classes in the IoT sensing task. 
\end{itemize}
The primary goal is to effectively (1) detect data sample of unseen classes $\vec{x}^u$ from $D^{test}$ fed to the specialist DNN $f(\cdot)$ on the local edge devices and then (2) leverage the cloud's FM $\Phi(\cdot)$ to perform zero-shot classification by assigning correct label $y^u \in U$ for the detected data of unseen classes. 
Note that an alternative way is to upload all the data to the cloud's FM for classification. However, in \sect \ref{ablation_study}, we will demonstrate that having the detection step alleviates the GZSL biasing problem. 
Such a cooperative system is common in IoT applications such as healthcare monitoring, autonomous driving, and AR/VR gaming. 
To achieve the goal, given an incoming IoT data sample, we first extract its IoT embedding and conduct open-set detection to determine whether the sample belongs to a seen class or unseen class, both on the edge. If it is detected as a seen class sample, we use the local specialist DNN to give prediction. Otherwise, if the sample is considered as unseen class data, we upload it to the cloud's FM for zero-shot learning.

%% file: method.tex
\section{Methodology}
\label{sec:method}

The overview of our approach is shown in Fig.~\ref{fig:overview}, which consists of the class prototype extraction, IoT embedding extraction, model training, and zero-shot classification modules.

\subsection{Class Prototype Extraction}\label{sec_4_1}
In ZSL, \textit{class prototypes} encapsulate the essential characteristics of each class in the semantic space. During inference, the similarity between the data embedding and each class prototype is measured to determine the sample's class. In this work, we utilize the text encoder of the vision-language FM, CLIP  \cite{radford2021learning}, to extract class prototypes from task-specific hints, namely \textit{prompt}. Prompt can be engineered in the form of \textit{hard prompt}, which is natural language instructions, or \textit{soft prompt}, which is continuous, learnable vector representations. 
The hard prompt can integrate domain expert knowledge but needs to be manually engineered. The soft prompt can be automatically fine-tuned to adapt to various tasks but is not human-interpretable. To combine the advantages of both, we propose to use cross-attention to fuse the soft and hard prompts to generate effective and comprehensive class prototypes.

\textbf{Learnable Soft Prompt}.
The default prompt in CLIP is constructed by plugging the class name into a pre-defined prompt template, i.e., ``a photo of \{class name\}''. However, such a fixed prompt is difficult to adapt to downstream tasks. Because CLIP's default prompts tend to gather together in the semantic space, which is unfavorable for data-text alignment \cite{zhou2022learning}.
To address this and avoid laborious manual prompt engineering, we learn a soft prompt end-to-end from training data, 
aiming to align the text embedding with IoT data embedding.
We follow the work in \cite{zhou2022learning} and place the class token in the middle of the prompt. For each class $c$, the learnable soft prompt fed to the pre-trained CLIP's text encoder $\Phi_{\text {text}}(\cdot)$ is represented by $\vec{p}^l(c) = \oplus (\vec{l}_1, ..., {\text {CLIPtokenizer}}[c], ..., \vec{l}_M)$,
where $\oplus$ is the concatenation operation, $\vec{l}_i$, $(i = 1 \cdots M)$ denotes the $i$-th learnable token vector, and $c$ is the class name, e.g., ``walking forward''. 
The learnable prompt is optimized over the training data 
using the loss defined shortly in Eq.~\ref{eq:contrastive-loss}.
The extracted learnable text embedding $\vec{t}^l(c) = \Phi_{\text {text}}(\vec{p}^l(c))$ has the same dimension as the IoT data embedding. 
The learned prompt token vectors $\vec{l}_i$, $(i = 1 \cdots M)$ are shared for all classes, which are task-specific.

\begin{figure*}
    \centering
    \includegraphics[width=\textwidth]{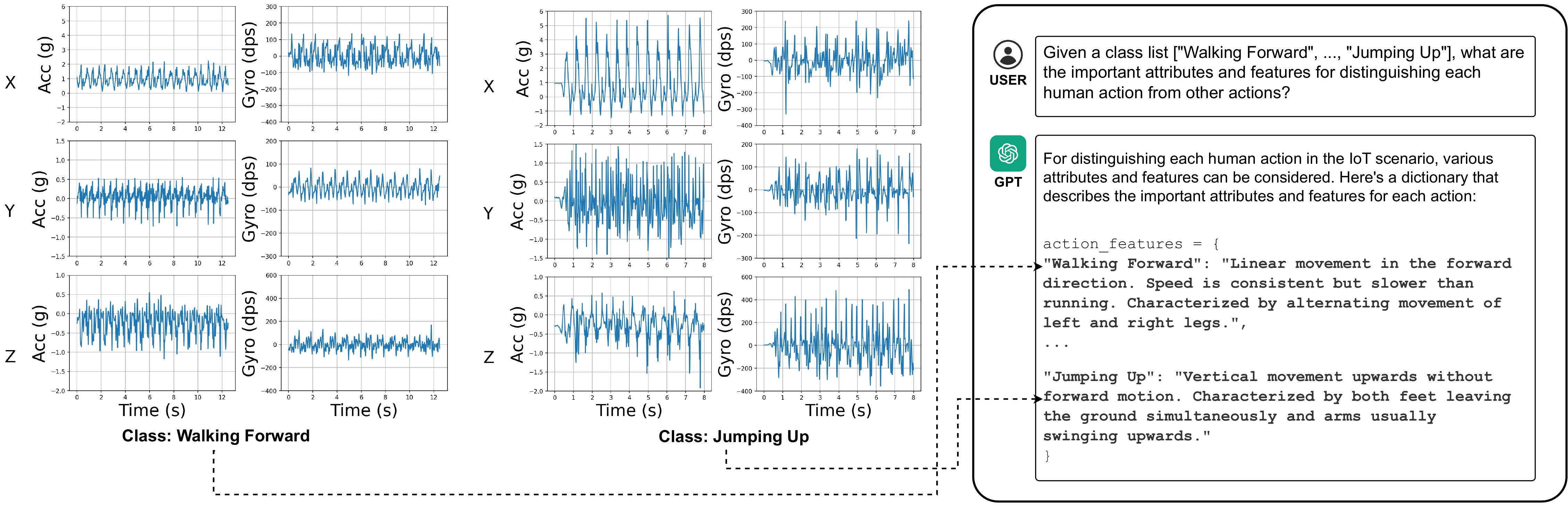}
    \caption{Visualization of two data samples from an IMU activity recognition dataset \cite{zhang2012usc}. X, Y, and Z axes are aligned with gravity, walking direction, and perpendicular to walking direction, respectively. The data sample of class ``walking forward'' has around zero values in the Y-axis of the accelerometer reading, indicating a constant speed along the walking direction. The data sample of ``jumping up'' has large positive values in the X-axis of the accelerometer reading, indicating vertical movements upwards. The patterns of the samples are characterized by the generated descriptive text.}
    \label{fig:auxiliary-hard-prompt}
    \vspace{1em}
\end{figure*}

\textbf{Auxiliary Hard Prompt.}
The learnable soft prompt provides task-specific context by aligning the text embedding with the IoT data embedding in the semantic space. Meanwhile, IoT data is usually characterized by certain physics principles, which can be leveraged as a strong supervision for prompt crafting. For example, Fig.~\ref{fig:auxiliary-hard-prompt} shows that the data samples of two classes in the USC-HAD \cite{zhang2012usc} dataset exhibit different patterns, which can be utilized to easily distinguish the data of the two classes. To leverage the physics principles governing the generation of the IoT sensor signals, we further use a hard prompt to give auxiliary class-specific information for constructing semantic embeddings. To automate the process, we use a state-of-the-art large language model (LLM), GPT-3.5 \cite{brown2020language}, to generate class-conditional descriptive text and fine-tune the text manually. For an IoT sensing task, we first feed the list of all classes to the LLM. Then, for each class $c$, we query the LLM: ``What are the important attributes and features to distinguish class $c$ from all the other classes?''. We then tokenized the answer to derive the auxiliary hard prompt $\vec{p}^a(c)$, which will be fed to CLIP's text encoder to derive the auxiliary text embedding $\vec{t}^a(c) = \Phi_{\text {text}}(\vec{p}^a(c))$, which has the same dimension as the IoT embeddings. Fig.~\ref{fig:auxiliary-hard-prompt} shows some example answers generated by GPT.

\textbf{Cross-Attention for Combining Prompts}.
To leverage the advantages of both the learnable soft prompt and auxiliary hard prompt, we combine the text embeddings of the two prompts using the cross-attention \cite{chen2021crossvit}, which is an attention mechanism for fusing two different sequences. In particular, we set $\vec{t}^a$ as the key input, denoted by $\vec{K}$, and $\vec{t}^l$ as the query and value inputs, denoted by $\vec{Q}$ and $\vec{V}$, respectively. The idea is to compute the attention weights between the query and key inputs, which embed the useful class-specific context information from $\vec{t}^a$, and then use the weights to aggregate the value input $\vec{t}^l$. Specifically, $\vec{Q} = \rho_{\vec{Q}}(\vec{t}^l)$, $ \vec{K} = \rho_{\vec{K}}(\vec{t}^a)$, and $\vec{V} = \rho_{\vec{V}}(\vec{t}^l)$, where $\rho_m(\cdot), (m \in \{\vec{Q}, \vec{K}, \vec{V}\}$) is a single-layer fully-connected neural network. $\rho_m(\cdot)$ is optimized over the training data on the loss defined shortly in Eq.~\ref{eq:contrastive-loss}.
The attention weights are computed by $\vec{A} = \text {softmax}(\frac{\vec{Q} \vec{K}^T}{\sqrt{d_\vec{K}}})$, where $d_\vec{K}$ is the dimension of $\vec{K}$. The output embedding, denoted by $\vec{t} = \vec{A}\vec{V}$, is the class prototype.
The semantic space is formed by the set of all class prototypes $\mathcal{T} = \{\vec{t}(c) \mid c \in \mathcal{S} \cup \mathcal{U}\}$.

\subsection{IoT Embedding Extraction}\label{sec_4_2}
For each input IoT data $\vec{x}_i$, we first use a feature extractor $\mu(\cdot)$ to extract its features $\vec{h}_i = \mu(\vec{x}_i)$. The feature extractor $\mu(\cdot)$ can be a commonly-used encoder like CNN, ResNet, and Transformer, which is decided by the IoT sensing modality. 
Then, we use an embedding projector $g(\cdot)$ aiming to project the IoT features $\vec{h}_i$ into {\blue the shared semantic space for alignment with class prototypes} and derive the IoT embeddings $\vec{e}_i = g(\vec{h}_i)$.
\subsection{Model Training}\label{sec_4_3}
We freeze the text encoder of CLIP $\Phi_{\text {text}}(\cdot)$ and conduct model training under the supervised contrastive learning strategy, which trains the models to distinguish between similar (positive) and dissimilar (negative) data sample pairs. This allows us to learn effective representations by maximizing the distance between different classes and minimizing the distance within the same class \cite{khosla2020supervised}.

\textbf{Supervised Contrastive Learning}. First, we jointly train the learnable soft prompt $\vec{p}^l$, $\rho_k(\cdot)$ in the cross-attention module, IoT feature extractor $\mu(\cdot)$, and IoT embedding projector $g(\cdot)$ on the seen train set $D^s$ using a supervised contrastive loss.
Within a batch of randomly sampled data $\{(\vec{x}^s_i, y^s_i)\}^{N_B}_{i = 1}$ from $D^s$, the positive pairs contain (1) two IoT data samples belonging to the same class and (2) an IoT data sample and its class label text. The negative pairs consist of (1) two IoT data samples belonging to different classes; (2) an IoT data sample and a class label other than its own; and (3) two different class labels.
The loss pulls together embeddings of positive pairs while pushing away the embeddings of negative pairs.
Let $i \in I \equiv \{1 \cdots N_B\}$ be the index of the data in a train batch. Let $N_T$ represent the number of distinct classes in the batch and $j \in J \equiv \{1 \cdots N_T\}$ be the index of distinct classes.
We define the supervised contrastive loss as:
\begin{equation}
\label{eq:contrastive-loss}
\begin{split}
\mathcal{L} &= \sum_{i \in I} \mathcal{L}_{i} \\
&= \sum_{i \in I} \bigg(\frac{-1}{|P(i)+1|} \cdot \bigg( \sum_{p \in P(i)} {\vec{e}_i \cdot \vec{e}_p}/{\tau} + {\vec{e}_i \cdot \vec{t}_j}/{\tau} \bigg) \\
&\quad + \log \bigg( \sum_{a \in A(i)} \exp \left({\vec{e}_i \cdot \vec{e}_a}/{\tau}\right) \\
&\quad + \sum_{n \in N(j)} \left( \exp \left({\vec{e}_i \cdot \vec{t}_n}/{\tau}\right) + \exp \left({\vec{t}_j \cdot \vec{t}_n}/{\tau}\right) \right) \bigg) \bigg),
\end{split}
\end{equation}
where, for each IoT data sample $\vec{x}_i$, $\vec{e}_i$ is its IoT embedding, $\vec{t}_j$ is its corresponding class prototype, $A(i) \equiv I \backslash \{i\}$, $N(j) \equiv J \backslash \{j\}$, $P(i) \equiv \{p \in A(i) : y_p = y_i\}$, and $\tau$ is a positive temperature scalar.

\textbf{Data Augmentation for Fine-Tuning}. During the model training on $D^s$ described in the previous paragraph, the IoT feature extractor and embedding projector are only trained on data of seen classes. Consequently, 
the IoT embeddings of unseen classes are biased to the seen ones, and thus, the data samples of unseen classes may easily be classified as seen ones. To address this bias problem, we propose to train a generative model under the Generative Adversarial Network (GAN) setting to synthesize data samples of unseen classes. The goal is to derive more robust IoT embeddings by fine-tuning the IoT feature extractor and embedding projector using the augmented unseen class data.
Given the train set $D^s$, we learn a conditional generator $G(\cdot)$ that takes as input the class prototype $\vec{t}(y)$ and a random Gaussian noise vector $\vec{z}$, aiming to output the synthesized IoT data $\tilde{\vec{x}} \in \mathcal{X}$ of class $y$. Note that the class prototype $\vec{t}(y)$ is generated by the frozen text branch. 
To achieve this goal, we modify the loss in \cite{xian2018feature} and define the data augmentation loss as: $\mathcal{L}_\text{DA} = \mathcal{L}_\text{WGAN}+ \mathcal{L}_\text{CLS}$. Specifically, $\mathcal{L}_\text{WGAN} = \mathbb{E}[D(\vec{x}, \vec{t}(y))]-\mathbb{E}[D(\tilde{\vec{x}}, \vec{t}(y))]-\xi \mathbb{E}\left[\left(\left\|\nabla_{\hat{\vec{x}}} D(\hat{\vec{x}}, \vec{t}(y))\right\|_2-1\right)^2\right]$, where $D(\cdot)$ is the discriminator, $\vec{x}$ is the real data, $\tilde{\vec{x}}=G(\vec{z},\vec{t})$ is the generated data, $\hat{\vec{x}} = \alpha\vec{x}+(1-\alpha)\tilde{\vec{x}}$ with $\alpha \sim U(0,1)$, and $\xi$ is the penalty coefficient. $\mathcal{L}_\text{CLS} = -\mathbb{E}[\log \text{Pr}(y \mid \tilde{\vec{x}} ; \theta)]$ is the classification loss computed by a linear softmax classifier parameterized by $\theta$ that is pre-trained on $D^s$.
The generator is trained by optimizing the objective: $\min _G \max _D \mathcal{L}_\text{DA}$.
The generator $G(\cdot)$ aims to fool the discriminator $D(\cdot)$ by generating IoT data that are considered as real, while the discriminator $D(\cdot)$ aims to distinguish real data from the synthesized one.
After $G(\cdot)$ is trained, we use it to generate a synthesized train set of unseen classes $D^{aug} = \{(\tilde{\vec{x}}^u_i,y^u_i)\}^{N_{{aug}}}_{i = 1} \in \mathcal{X} \times \mathcal{U}$ and use it to fine-tune the IoT feature extractor and embedding projector using the loss defined in Eq.~\ref{eq:contrastive-loss}.

\subsection{Zero-Shot Classification}\label{sec_4_4}
As outlined in \sect\ref{sec:problem-formulation}, we decompose the zero-shot classification into two steps. The first step is to identify unseen class data, i.e., \textit{open-set detection}, on the local edge devices. The second step is to conduct \textit{zero-shot learning} using the FM located on the cloud. 

\textbf{Open-Set Detection} is a binary classification problem to identify whether a data sample belongs to seen or unseen classes. Inspired by the work in \cite{sun2022out}, we develop a distance-based method for open-set detection. 
First, based on a train set $D^s$, we cluster the IoT embeddings of all data samples based on their classes and denote these clusters by $\{E^s_i\}^{N_s}_{i=1}$, where $N_s$ is the number of seen classes. Each class cluster $E^s_i$, $(i=1\cdots N_s)$ consists of a set of IoT embeddings $\{\vec{e}_{i,j}\}_{j=1}^{N_i}$, where $N_i$ is the number of data samples in $E^s_i$.
For an input data sample $\vec{x}^\text{test} \in D^{test}$, which may belong to either seen or unseen classes, we compute the Euclidean distances between its IoT embedding $\vec{e}^{\text{test}}$ and the IoT embeddings in each class cluster $E^s_i$ as $d_{i,j} = \Vert \vec{e}^{\text{test}} - \vec{e}_{i,j} \Vert_2, \vec{e}_{i,j} \in E^s_i$. We sort $d_{i,j}$ to obtain the $k_i$-th smallest distance for each cluster, denoted by $d_i^{(k_i)}$. We use a simple threshold-based criterion on $d_i^{(k_i)}$ to determine whether the input sample belongs to seen or unseen classes:
\begin{gather}
    Q(\vec{x}^\text{test};k_i)=\sum_i^{N_s}\mathbf{1}|d_i^{(k_i)} \le \lambda_i |,\\
    S_{\text {open}}(\vec{x}^\text{test}) =
\begin{cases}
{\rm Unseen}, Q=0 \\
{\rm Seen}, Q \geq 1 \\
\end{cases},
\end{gather}
where $\mathbf{1}|\cdot|$ is the indicator function and $\lambda_i$ is the class-specific distance threshold that is decided empirically by correctly associating a high
fraction of seen class data samples to their corresponding class clusters using a validation set. If the value of $Q$ equals 0, it indicates that the test sample does not belong to any seen class clusters and should be considered as unseen. If the value of $Q \geq 1$, it means that the test sample can be associated with at least one seen class cluster and should be considered as seen.

\textbf{Zero-Shot Learning}.
For a detected ``unseen'' test sample $\vec{x}^{\text{det}}$ with IoT embedding $\vec{e}^{\text{det}}$, we upload it to the cloud's FM for zero-shot learning. Specifically, we compute the similarity scores, i.e., dot product, between $\vec{e}^{\text{det}}$ and all the class prototypes in $\{\vec{t}(c_i^u), c_i^u \in U\}$. Then, the class with the highest similarity score is the predicted label $\hat{y}^{\text{det}}$ for $\vec{x}^{\text{det}}$:
{\blue
\begin{gather}
    \hat{y}^{\text{det}} = \mathop{\rm{argmax}}\limits_{c_i^u \in U}(\vec{e}^{\text{det}}\cdot \vec{t}(c_i^u)^T),
\end{gather}
}

%% file: eval.tex
\section{Evaluation}
\label{sec:eval}

\subsection{Datasets}
{\blue We evaluate our approach on multiple IMU, mmWave, and Wi-Fi datasets that are commonly used in IoT sensing tasks as follows. 
}

\textbf{USC-HAD} \cite{zhang2012usc}. The USC Human Activity Dataset is an IMU dataset of 12 different daily activities collected from 14 human subjects. By sampling it with a 1.28-second window and a 50\% overlap rate, we obtain 42,708 samples, each consisting of 1.28-second 3-axis accelerometer and 3-axis gyroscope readings. We divide the activities into 9 seen classes and 3 unseen classes.

\textbf{PAMAP2} \cite{reiss2012introducing}. 
The Physical Activity Monitoring Dataset consists of 12 daily activities by collecting IMU data following a protocol from 9 subjects. We divide the activities into 9 seen classes and 3 unseen classes. We adopt a 1.71-second sliding window with a 10\% overlap rate to 
extract 4,178 samples.

\textbf{MM-Fi} \cite{yang2024mm}. 
The MM-Fi dataset is a multi-modal wireless human sensing dataset consisting of 1,080 consecutive sequences with over 320k synchronized frames from five sensing
modalities.
We adopt the Wi-Fi and filtered mmWave sub-datasets in environment 4 from the MM-Fi. We resample mmWave and Wi-Fi data using 1-second and 0.6-second sliding windows with 10\% overlap, respectively, yielding 27,337 mmWave samples and 8,748 Wi-Fi samples. For both mmWave and Wi-Fi, we split the 27 activity classes into 22 seen classes and 5 unseen classes.

We adopt a $K$-fold evaluation strategy to split each dataset into seen classes and unseen classes. For USC-HAD and PAMAP2, we randomly select 3 unseen classes in each of $K$=4 folds.
For mmWave and Wi-Fi, we randomly select 5 unseen classes in $K$=5 folds.
For the seen class data samples, we divide them into training, validation, and test sets with a ratio of 8:1:1. The validation set is used to tune the parameters like $\lambda_{i}$. The test set has equal number of seen class and unseen class data samples.

\subsection{Implementation Details}
We use Pytorch to implement our approach.
We use Vision Transformer as the IoT Feature Extractor for all modalities. For class prototype extraction, we use GPT-3.5 to generate auxiliary hard prompts. The text encoder is adopted from the frozen CLIP text encoder with ViT-B/16 backbone. The supervised contrastive loss's temperature parameter $\tau$ is set to 0.2. For data augmentation, the random Gaussian noise vector $\vec{z}$ follows a normal distribution $\mathcal{N} \sim(0,1)$, and the penalty coefficient $\xi$ is set to 10. During training, the optimization is performed via the Stochastic Gradient Descent with Momentum (SGDM) algorithm. The learning rate is 0.001 and the batch size for training is 64. In open-set detection, the $k_i$ is set to $0.08\times N_i$. The threshold $\lambda_i$ is set to a number that guarantees a large percentage of seen data in validation set can be successfully classified. This percentage is set to 80\% for USC-HAD, PAMAP2, MM-Fi (Wi-Fi), and 75\% for MM-Fi (mmWave). All results are obtained by calculating the mean and variance on all splits for each dataset.

\subsection{Open-Set Detection Performance} 

\subsubsection{Baselines and Evaluation Metrics}

We consider the following open-set detection baselines.

\textbf{MSP} \cite{hendrycks2016baseline} measures the maximum softmax probability generated by a model trained on the seen class data using cross-entropy loss to detect unseen class data. 
For the MSP baseline, we adopt the Vision Transformer as model architecture.

\textbf{KNN} \cite{sun2022out} computes the $k$-th nearest neighbor distance between
an input image feature and the training set for unseen class data detection. The images are augmented, e.g., by adding Gaussian noise, for supervised contrastive learning in KNN. 
In the KNN baseline, we augment the IoT data also by adding noise and use supervised contrastive learning to extract IoT embeddings. 

\textbf{MCM} \cite{ming2022delving} measures the distance between an input image feature and its closest label embedding, both directly generated by a large vision-language FM, for unseen class data detection. For the MCM baseline, we replace the image features with our IoT embeddings and use the prompt template "The human action of [CLASS]" for text encoding.

For all open-set detection baselines, we set the detection thresholds and parameters using the same strategy as our method.

To evaluate the performance of open-set detection, we employ the weighted precision, recall, and F1 score.

\subsubsection{Results}
In Table~\ref{tab:open_set}, we can see that our approach achieves the best open-set detection performance compared with all baselines on all three modalities' datasets. In detail, our approach outperforms the traditional softmax-based method MSP because the supervised contrastive loss can help our model obtain more distinguishable IoT embeddings than the cross-entropy loss in MSP. The KNN method performs worse than ours. This is because the image augmentation used by KNN for supervised contrastive learning, e.g., adding noise, cannot be directly applied to IoT data. Differently, our approach aligns text embeddings with IoT embeddings using supervised contrastive learning and achieves more generalized IoT embeddings. Our approach performs better than MCM since the MCM only takes hard prompts to generate text embeddings, which is undesirable for aligning IoT embeddings of different tasks with text embeddings.

\begin{table}[]
\centering
\begin{tabular}{@{}ccccc@{}}
\toprule
\multirow{2}{*}{Dataset} & \multirow{2}{*}{Method} & \multicolumn{3}{c}{Performance} \\ \cmidrule(l){3-5} 
                         &                         & Precision  & Recall  & F1 score \\ \cmidrule(r){1-5}
\multirow{4}{*}{\parbox[c]{4em}{\centering MM-Fi\\(mmWave)}}  & MSP                     & 72.1±0.1\% & 71.9±0.1\%& 71.8±0.1\%   \\
                         & KNN  & 68.9±0.0\% & 68.5±0.1\% & 68.4±0.1\%         \\
                         & MCM                     & 70.8±0.2\% & 70.5±0.3\% & 70.4±0.3\%  \\
                         & Ours               & \textbf{73.5}±0.1\% & \textbf{73.2}±0.1\% & \textbf{73.0}±0.1\%       \\ \cmidrule(r){1-5}
\multirow{4}{*}{USC-HAD}  & MSP    & 69.4±0.3\%  & 68.6±0.4\% & 67.8±0.6\% \\
        & KNN    & 77.8±0.1\%  & 77.7±0.1\% & 77.7±0.1\% \\
        & MCM    & 66.8±1.2\%  & 65.7±1.3\% & 64.1±1.7\% \\
        & Ours   & \textbf{79.2}±0.3\%  & \textbf{78.9}±0.3\% & \textbf{78.8}±0.3\%          \\ \cmidrule(r){1-5}
\multirow{4}{*}{PAMAP2}   & MSP    & 87.6±0.1\%  & 87.0±0.0\% & 87.0±0.0\% \\
        & KNN    & 88.7±0.1\%  & 87.7±0.1\% & 87.6±0.1\% \\
        & MCM    & 81.4±0.3\%  & 81.1±0.2\% & 81.1±0.2\% \\
        & Ours   & \textbf{89.6}±0.0\%  & \textbf{88.0}±0.0\% & \textbf{87.9}±0.0\%          \\ \cmidrule(r){1-5}
\multirow{4}{*}{\parbox[c]{4em}{\centering MM-Fi\\(Wi-Fi)}}   & MSP                & 77.2±0.1\%          & 77.0±0.1\%          & 77.0±0.1\%     \\
                         & KNN                    & 58.1±0.1\%           & 56.5±0.1\%           & 54.0±0.1\%          \\
                         & MCM                    & 74.0±0.1\%          & 73.6±0.1\%          & 73.4±0.1\%       \\
                         & Ours                  & \textbf{77.4}±0.0\%          & \textbf{77.3}±0.0\%          & \textbf{77.3}±0.0\%       \\ \bottomrule                         
\end{tabular}
    \caption{Open-set detection performance.}
    \label{tab:open_set}
\end{table}

\subsection{Zero-Shot Classification Performance}
\subsubsection{Baselines and Evaluation Metrics}

We consider the following baselines for evaluating the GZSL performance of our approach.

\textbf{ALE} \cite{akata2015label} measures the compatibility of image features and class label embeddings in the Euclidean space for ZSL. 

\textbf{DCN} \cite{liu2018generalized} uses a Deep Calibration Network to map image features and class prototypes to a common embedding space for ZSL.

\textbf{BERT} \cite{devlin2018bert} We replace the frozen CLIP text encoder in our approach with the pre-trained BERT to process the prompt template as a baseline.

\textbf{f-CLSWGAN} \cite{xian2018feature} 
uses an attribute conditional feature generating adversarial network to generate CNN features of unseen classes for ZSL.

\textbf{FREE} \cite{chen2021free} learns a visual feature generator jointly with a feature refinement module for ZSL.

ALE, DCN, and BERT are embedding-based methods, while f-CLSWGAN and FREE are generative-based methods. We replace the image features with IoT embeddings in all the above methods as baselines.

We evaluate the performance of GZSL using the following metrics. We measure the percentage of correctly classified seen and unseen class data samples, i.e., seen class accuracy $\text{ACC}_{\mathcal{S}}$ and unseen class accuracy $\text{ACC}_{\mathcal{U}}$, respectively. 
Note that these accuracies are the weighted average across all seen/unseen classes.
We also compute the harmonic mean \cite{pourpanah2022review}, which is a conventional metric to measure the inherent biasness of a GZSL method with respect to the seen classes: 
\begin{gather}
    \text{ACC}_H = \frac{2\times \text{ACC}_{\mathcal{S}} \times \text{ACC}_\mathcal{U}}{\text{ACC}_{\mathcal{S}} + \text{ACC}_\mathcal{U}},
\end{gather}
A lower $\text{ACC}_H$ means that the unseen class accuracy $\text{ACC}_{\mathcal{U}}$ is lower than seen class accuracy $\text{ACC}_{\mathcal{S}}$, indicating that a GZSL method is biased towards the seen classes.

\subsubsection{Results}
As shown in Table \ref{tab:gzsl}, our approach achieves the best $\text{ACC}_\mathcal{U}$ and $\text{ACC}_H$ on all datasets compared with all baselines. Although some baselines have higher $\text{ACC}_\mathcal{S}$, it is impractical to only consider seen classes since recognizing both seen and unseen classes is critical for most IoT sensing tasks.
Specifically, our approach outperforms embedding-based approaches ALE, DCN, and BERT on $\text{ACC}_H$ because we construct better text embeddings by using cross-attention to integrate soft prompt and hard prompt while using contrastive loss to make text-IoT embedding alignment more accurate and robust. Moreover, these methods are trained only with uni-modal textual data, whereas the CLIP text encoder is trained from multi-modal data of both images and text, which generates more effective text embeddings for data-text alignment \cite{chen2023difference}. Compared with generative methods f-CLSWAGAN and FREE, our approach still achieves superior performance. The generative methods' results on small-scale IoT datasets are less satisfactory because their performance relies on a large amount of training data. For our approach, in addition to using the generative model for synthesizing unseen class data to alleviate the biasing problem, the open-set detection also helps our method further classify the seen and unseen data correctly.

\begin{table}[ht!]
    \centering
\begin{tabular}{@{}ccccc@{}}
\toprule
\multirow{2}{*}{Dataset} & \multirow{2}{*}{Method} & \multicolumn{3}{c}{Performance} \\ \cmidrule(l){3-5} 
                         &                         & $\text{ACC}_\mathcal{S}$  & $\text{ACC}_\mathcal{U}$  & $\text {ACC}_H$ \\ \cmidrule(r){1-5}
\multirow{4}{*}{\parbox[c]{4em}{\centering MM-Fi\\(mmWave)}}  & ALE                      & 86.5±0.1\%           & 0.01±0.0\%           & 2.0±0.0\%    \\
                         & DCN                      & 67.0±1.3\%           & 30.2±1.3\%           & 40.3±0.9\%     \\
                         & BERT & 71.8±0.0\% & 36.9±0.6\% & 48.3±0.5\% \\                         
                         & f-CLSWGAN                & 77.2±0.3\%           & 29.7±0.5\%           & 42.3±0.4\%     \\
                         & FREE                   & 87.7±0.1\%  & 25.3±0.8\%           & 38.3±1.1\%     \\
                         & Ours                    & 73.3±0.0\%           & 40.4±0.5\%  & \textbf{51.7}±0.3\%    \\ \cmidrule(r){1-5}
\multirow{4}{*}{USC-HAD}   & ALE                    & 92.5±0.0\%  & 0.6±0.0\%            & 1.1±0.0\%   \\
                         & DCN                     & 56.6±3.2\%           & 37.1±1.5\%           & 43.3±1.1\%        \\
                         & BERT & 74.9±0.1\% & 41.6±1.3\% & 52.2±0.7\% \\                         
                         & f-CLSWGAN                    & 81.3±0.5\%           & 29.2±3.5\%           & 39.5±4.9\%           \\
                         & FREE                     & 90.9±0.1\%           & 14.0±0.6\%           & 23.2±1.6\%         \\                         
                         & Ours                   & 73.1±0.5\%           & 54.8±1.8\%  & \textbf{61.1}±0.7\%    \\ \cmidrule(r){1-5}
\multirow{4}{*}{PAMAP2}   & ALE                     & 70.1±3.9\%  & 12.1±1.9\%            & 15.5±3.6\%   \\
                         & DCN                      & 42.2±0.9\%           & 33.1±0.2\%           & 36.7±0.3\%         \\
                         & BERT & 74.7±0.0\%  &  49.9±0.7\% & 59.3±0.0\% \\                          
                         & f-CLSWGAN                    & 92.4±0.2\%           & 27.8±1.1\%           & 41.7±1.5\%         \\
                         & FREE                    & 87.7±0.3\%           & 37.2±0.2\%           & 52.1±0.2\%        \\                       
                         & Ours                    &    74.6±0.1\%                       &   53.7±0.4\%                         &    \textbf{62.1}±0.2\%  \\ \cmidrule(r){1-5}
\multirow{4}{*}{\parbox[c]{4em}{\centering MM-Fi\\(Wi-Fi)}}   & ALE                      & 52.2±6.0\%           & 9.5±0.5\%           & 11.8±0.5\%     \\
                         & DCN                     & 60.1±1.8\%           & 18.7±0.2\%           & 28.2±0.4\%         \\
                         & BERT & 62.5±0.0\% & 29.5±0.5\% & 39.5±0.5\% \\                               
                         & f-CLSWGAN                    & 84.7±0.0\%  & 6.2±0.1\%            & 11.4±0.1\%          \\
                         & FREE                   & 80.0±0.1\%          & 30.4±0.3\%           & 43.6±0.3\%      \\                   
                         & Ours                   & 75.1±0.0\%                         &   35.3±0.5\%                         &     \textbf{47.6}±0.4\%         \\ \bottomrule                               
\end{tabular}
    \caption{Generalized zero-shot learning performance.}
    \label{tab:gzsl}
\end{table}

\subsection{Ablation Study}\label{ablation_study}
To analyze the effectiveness of the prompt engineering, open-set detection, and data augmentation modules, we conduct ablation studies to remove one of the components. The results are shown in Table \ref{tab:ablation}.

\textbf{Prompt Engineering}. To demonstrate that prompt engineering brings improvement to GZSL, we remove it by replacing the prompt engineering part with a fixed prompt template, "The human action of [CLASS]". As shown in Table \ref{tab:ablation}, we can see that there is an accuracy drop in $\text{ACC}_{\mathcal{U}}$ and $\text{ACC}_{H}$  by disabling the prompt engineering.
The prompt engineering provides tailored text embeddings by integrating the soft prompt and hard prompt, helping the model to align text embeddings and IoT embeddings, resulting in better GZSL results.

\textbf{Open-Set Dectection}. To validate the effectiveness of open-set detection, we remove it and directly match the IoT embeddings with all seen and unseen text embeddings. The class label with the largest matching score will be the classification result. As shown in Table \ref{tab:ablation}, although there is an increase for $\text{ACC}_{\mathcal{S}}$, the $\text{ACC}_{\mathcal{U}}$ and $\text{ACC}_{H}$ experience a huge decline by removing the open-set detection. This is because the open-set detection helps the model eliminate the bias problem in GZSL, leading to classifying more unseen data correctly. 

\textbf{Data Augmentation}. To investigate the effectiveness of data augmentation, we remove the step of fine-tuning the model using synthetic data. From Table \ref{tab:ablation}, we can observe that by using data augmentation, the $\text{ACC}_{\mathcal{U}}$ is improved since the synthetic unseen data helps the model to reduce the bias problem of unseen IoT embeddings.

\begin{table}[]
\begin{tabular}{@{}ccccccc@{}}
\toprule
\multirow{2}{*}{Dataset} & \multirow{2}{*}{P.E.} & \multirow{2}{*}{O.S.} & \multirow{2}{*}{D.A.} & \multicolumn{3}{c}{Performance}                                                      \\ \cmidrule(l){5-7} 
                         &                       &                       &                       & $\text{ACC}_{\mathcal{S}}$ & $\text{ACC}_{\mathcal{U}}$ & $\text {ACC}_H$ \\ \cmidrule(r){1-7}
\multirow{4}{*}{\parbox[c]{4em}{\centering MM-Fi\\(mmWave)}}  &                       & \cmark & \cmark &   60.0±0.1\%                         &  38.1±0.4\%                          &        46.4±0.2\%                    \\
                         & \cmark &                       & \cmark & 88.6±0.0\%                           & 8.4±0.1\%                            & 15.0±0.5\%                            \\
                         & \cmark & \cmark &                       & 72.0±0.0\%                            &  39.8±0.2\%                          &51.1±0.1\%                            \\
                         & \cmark & \cmark & \cmark &   73.3±0.0\%                         &40.4±0.5\%                            & \textbf{51.7}±0.3\%                            \\ \cmidrule(r){1-7}
\multirow{4}{*}{USC-HAD}   &                       & \cmark & \cmark &   74.8±0.2\%                         & 40.3±3.0\%                            &   49.9±1.5\%                         \\
                         & \cmark &                       & \cmark &  83.8±1.1\%                          & 14.1±0.9\%                           &     22.8±1.7\%                       \\
                         & \cmark & \cmark &                       &       73.1±0.3\%                     &     51.3±1.1\%                       &   59.5±0.5\%                         \\ 
                         & \cmark & \cmark & \cmark &      73.1±0.5\%                &        54.8±1.8\%                    &    \textbf{61.1}±0.7\%                        \\ \cmidrule(r){1-7}
\multirow{4}{*}{PAMAP2}   &                       & \cmark & \cmark &         73.5±0.1\%                   &  53.1±0.9\%                          &      61.2±0.4\%                      \\
                         & \cmark &                       & \cmark &    92.9±0.2\%                        &             7.7±0.9\%               &  12.9±2.1\%                          \\
                         & \cmark & \cmark &                       &    74.7±0.1\%                        &   52.7±0.2\%                         &           61.6±0.0\%                 \\
                         & \cmark & \cmark & \cmark &    74.6±0.1\%                       &   53.7±0.4\%                         &    \textbf{62.1}±0.2\%                        \\ \cmidrule(r){1-7}
\multirow{4}{*}{\parbox[c]{4em}{\centering MM-Fi\\(Wi-Fi)}}   &                       & \cmark & \cmark &             65.6±0.1\%               &      31.6±0.3\%                      &      42.1±0.2\%                      \\
                         & \cmark &                       & \cmark &  80.0±0.2\%                          &   10.2±0.1\%                         &  17.7±0.5\%                          \\
                         & \cmark & \cmark &                       &   74.8±0.0\%                       &     34.4±0.4\%                         &    46.7±0.4\%                          \\
                         & \cmark & \cmark & \cmark &  75.1±0.0\%                         &   35.3±0.5\%                         &     \textbf{47.6}±0.4\%                       \\ \bottomrule  
\end{tabular}
    \caption{Ablation study. P.E. indicates prompt engineering, O.S. represents open-set detection, and D.A. is the data augmentation.}
    \label{tab:ablation}
\end{table}

%% file: conclude.tex
\section{Conclusion}
\label{sec:conclude}

In this work, we have explored the potential of foundation models (FMs) for zero-shot IoT sensing. We leverage the generalized knowledge encoded in FMs and employ novel techniques to bridge the semantic gap between IoT data and text embeddings. Cross-attention is utilized for effective prompt engineering and data augmentation to mitigate bias. The evaluation has demonstrated the superior performance of our approach compared with existing baselines in both open-set detection and generalized zero-shot learning tasks across USC-HAD, PAMAP2, MM-Fi datasets of IMU, mmWave, Wi-Fi modalities. Future research includes exploring the integration of additional modalities and adaptability of our approach to different IoT sensors and applications. Besides, investigating the explainability and interpretability of FM-based zero-shot IoT sensing would be valuable for understanding the decision-making process.

%% file: appendix.tex
\section{Appendix}
\subsection{Soft Prompt Examples}
\begin{figure}[h]
    \centering
    \includegraphics[width=0.5\textwidth]{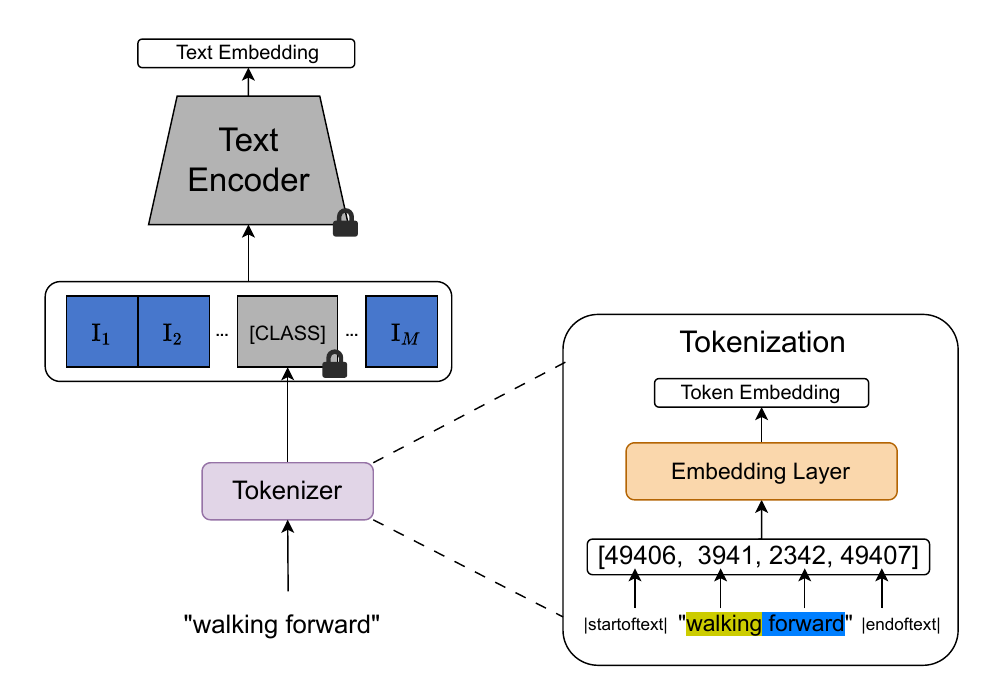}
    \caption{Soft learnable prompt optimization for CLIP~\cite{zhou2022learning}. The learnable context is composed of continuous vector $\vec{l}_i$ that can be optimized during learning. $[\mathrm{CLASS}]$ is the tokenized class label embedding. The parameters of $[\mathrm{CLASS}]$ and the text encoder are frozen during training.}
    \label{fig:soft_prompt}
\end{figure}

We provide an example of the learnable prompt for the text label ``walking forward" in this section. As shown in Fig.~\ref{fig:soft_prompt}, the input text ``walking forward" is first broken down into individual tokens, including the reserved start and end tokens. These tokens are then converted into token IDs, which are passed through the embedding layer to generate a corresponding token embedding. The token embedding is combined with learnable vectors to create a learnable prompt, which is then fed into the text encoder. We set the number of learnable vectors to be 8 for a balanced trade-off between the performance and generalizability of the learned prompts. Under this setting, the experiments in~\cite{zhou2022learning} show placing class tokens in the middle performs better than placing the tokens at the end of the prompts.

\subsection{Hard Prompt Examples}
The auxiliary hard prompts are used to encode IoT domain knowledge into the class prototypes. The hard prompts are generated as follows: First, we prepare a list of human action labels for each dataset. For example, the list of human action labels for USC-HAD~\cite{zhang2012usc} is [``Walking Forward'', ``Walking Left'', ``Walking Right'', ``Walking Upstairs'', ``Walking Downstairs'', ``Running Forward'', ``Jumping Up'', ``Sitting'', ``Standing'', ``Sleeping'', ``Elevator Up'', ``Elevator Down'']. Next, we place the action label list into the prompt template, as shown in Fig.~\ref{fig:auxiliary-hard-prompt}, to generate the auxiliary hard prompts. To reduce the manual effort in preparing the hard prompts, the large language models are required to return a key-value dictionary, where the original action label is the key and the generated description is the value. We present all the hard prompts of all datasets generated by the GPT-3.5 in our open-source code: https://github.com/schrodingho/FM\_ZSL\_IoT.

\subsection{Visualization}
\begin{figure}[h]
    \centering
    \includegraphics[width=0.5\textwidth]{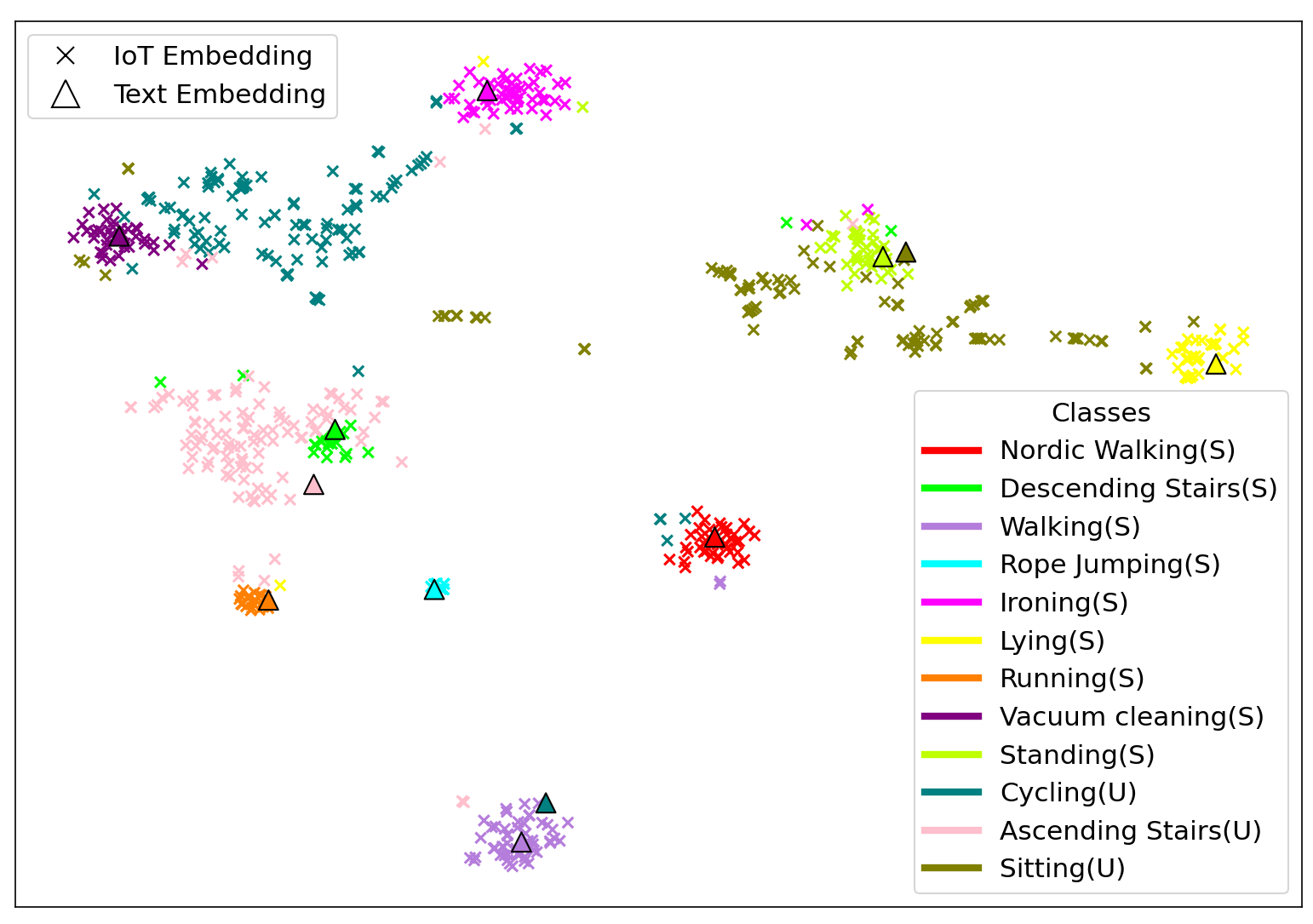}
    \caption{T-SNE visualization of PAMAP2~\cite{reiss2012introducing} testing data. The classes with (S) suffix are seen classes and the classes with (U) suffix are unseen classes.}
    \label{fig:tsne}
\end{figure}

Fig.~\ref{fig:tsne} presents the t-SNE visualization of PAMAP2 testing data's IoT embeddings and text embeddings in the unified embedding space. We can observe that the IoT and text embeddings of each seen class are closely aligned in the embedding space and different seen classes' embeddings can be distinguished. 
The supervised contrastive learning plays a vital role in the excellent alignment with a few IoT data samples. For unseen classes, the IoT embeddings of each unseen class can be distinguished from the seen classes, and each unseen class cluster has a relatively independent embedding space. However, unseen classes' IoT and text embeddings are not perfectly aligned, e.g. unseen text embedding ``Cycling'' and its corresponding IoT embeddings in Fig.~\ref{fig:tsne}. If we use all seen and unseen text embeddings for zero-shot classification, the unseen IoT embeddings can easily be misclassified as seen classes. To address this, we adopt the open-set detection to separate unseen and seen IoT embeddings. The separated unseen IoT embeddings are more likely to match correctly with unseen text embeddings.

